\documentclass[12pt]{article}
\usepackage[utf8]{inputenc}
\usepackage{xcolor}
\usepackage{float}
\usepackage[justification=centering]{caption}
\usepackage[margin=0.5in]{geometry}
\usepackage[pdftex]{graphicx}
\graphicspath{{\Documents}}     
\DeclareGraphicsExtensions{.pdf,.jpeg,.png}
\usepackage{titlesec}
\usepackage{moreverb}
\usepackage{makecell}
\usepackage[utf8]{inputenc}
\usepackage{multicol}
\usepackage{amsmath,amssymb,amsfonts}
\usepackage{algorithmic}
\usepackage{textcomp}
\usepackage{cite}
\usepackage{multirow}
\usepackage{float}
\usepackage{subcaption}
\usepackage[T1]{fontenc}
\usepackage[utf8]{inputenc}

\title{\textbf{Predictive Analysis for Optimizing Port Operations}}

\begin{multicols}{2}
\author{Aniruddha Rajendra Rao\\
Industrial AI Lab\\
Hitachi America, Ltd., R\&D\\
Santa Clara, USA\\
Aniruddha.Rao@hal.hitachi.com\\
\and
Haiyan Wang\\
Industrial AI Lab\\
Hitachi America, Ltd., R\&D\\
Santa Clara, USA\\
\vspace{1cm}
Haiyan.Wang@hal.hitachi.com\\
\and
Chetan Gupta\\
Industrial AI Lab\\
Hitachi America, Ltd., R\&D\\
Santa Clara, USA\\
\vspace{1cm}
Chetan.Gupta@hal.hitachi.com\\}
\end{multicols}

\date{}

\begin{document}

\maketitle
\vspace{-1.5cm}
\hspace{0.35cm}\textbf{Keywords:} Supervised Learning, Machine Learning, Feature Importance, Supply Chain, Port Analysis.

\abstract{Maritime transport is a pivotal logistics mode for the long-distance and bulk transportation of goods. However, the intricate planning involved in this mode is often hindered by uncertainties, including weather conditions, cargo diversity, and port dynamics, leading to increased costs. Consequently, accurate estimation of the total (stay) time of the vessel and any delays at the port are essential for efficient planning and scheduling of port operations. This study aims to develop predictive analytics to address the shortcomings in the previous works of port operations for a vessel’s Stay Time and Delay Time, offering a valuable contribution to the field of maritime logistics. The proposed solution is designed to assist decision-making in port environments and predict service delays. This is demonstrated through a case study on Brazil's ports. Additionally, feature analysis is used to understand the key factors impacting maritime logistics, enhancing the overall understanding of the complexities involved in port operations. Furthermore, we perform SHapley Additive exPlanations (SHAP) analysis to interpret the effects of the features on the outcomes and understand their impact on each sample, providing deeper insights into the factors influencing port operations.}

\section{Introduction}


In the intricate landscape of global maritime logistics, an often overlooked yet substantial segment of a vessel's journey unfolds within the confines of port facilities. While traditional narratives tend to focus on vessels (ships) navigating the open waters \cite{i1, i2, i3}, a substantial portion of their journey involves complex maneuvers, cargo handling, and operational intricacies within ports \cite{p1}. Surprisingly, this dualistic nature of a vessel's journey, at sea, and within port, the latter has received scant attention in the existing literature. The temporal dynamics, efficiency challenges, and optimization opportunities inherent in the port-centric leg of a vessel's journey warrant closer scrutiny \cite{brazildata}. This gap in scholarly exploration calls for an in-depth examination, shedding light on the pivotal role that port activities play in the broader context of maritime transportation, supply chain efficiency, and global trade networks \cite{p2}.

The optimization of port operations holds critical importance in the context of globalization, where nations strive to maintain their competitiveness and economic resilience in an interconnected world. Efficient port operations are essential not only for facilitating trade and commerce but also for accommodating the growing flow of goods and passengers associated with global tourism and cultural exchanges. The integration of marine traffic information through Application Programming Interface (API) \cite{Reggiannini2019RemoteSF, BautistaSnchez2019StatisticalAI, Eriksen2006MaritimeTM} has emerged as a key technological advancement, offering real-time insights into vessel movements and enabling proactive decision-making to enhance port efficiency on a global scale.

While the specific focus of this project is on Brazil's ports, the broader implications extend to a global context. Ports worldwide face similar challenges related to the optimization of operations, the impact of globalization on trade flows, and the integration of advanced technologies for enhanced efficiency \cite{p4, p5}. By studying the case of Brazil, with its rich maritime history and strategic positioning in global trade, valuable insights can be gained that are applicable to port operations and maritime logistics across diverse regions and economies.

The main contribution of the paper is as follows:
\begin{itemize}
    \item We define the problem for predictive analysis for port operations which is lacking in the current literature.
    \item We execute an exhaustive comparison of methods with multiple metrics for prediction and classification of Total Time (stay time) and Delay Time at the port.
    \item We discover key factors that affect port operations using feature importance.
    \item We perform SHAP (SHapley Additive exPlanations) analysis to understand the contribution of these key factors on the output.
\end{itemize}

The paper is structured as follows: Section 2 provides a comprehensive review of the literature relevant to the study. Section 3 discusses Brazil's port case study and outlines the proposed methods used for port predictive analysis. Section 4 dives into the results and shares key features which impact the port operations. We finish with Section 5, which offers the conclusions from the study, and suggests avenues for future research.

\section{Literature Review}

The Berth Allocation Problem (BAP) is a critical challenge in port operations, aiming to optimize the allocation of berthing facilities for the incoming ships while considering various restrictions and objectives \cite{Bierwirth2015AFS, h2}. The estimation of different durations for the vessels is a crucial factor in BAP mathematical formulations, as it directly influences berth allocation decisions. Therefore, the development of efficient prediction and classification models for Total Time is strategically important for addressing the BAP. Studies \cite{lr1, Rodrigues2022BerthAA} have explored novel algorithms and optimization techniques to address the BAP, emphasizing the importance of accurate Total Time predictions in optimizing berth allocations.

Predictive Model Scheduling (PMS) is another area of interest in port operations, focusing on the development of predictive models to optimize scheduling processes for port activities \cite{Dulebenets2018ACM, Boetzelaer2013MastersTM}. By leveraging predictive analytics for Total Time and Delay Time, PMS aims to enhance the efficiency and effectiveness of port operations by forecasting demand, optimizing resource allocation, and improving overall performance \cite{Weerasinghe2023OptimizingCT}. These works have explored the application of machine learning algorithms for predictive scheduling in container terminals, demonstrating the potential of these methods in improving port productivity.

Performance management of ports and vessels is a critical aspect of maritime logistics, involving the monitoring, analysis, and optimization of key performance indicators (KPIs) to ensure operational efficiency and competitiveness. \cite{Farag2020TheDO, p3} emphasizes the importance of performance management systems for ports, highlighting the need for real-time data analytics and decision support tools to enhance operational performance. Regarding port and ship characteristics related to faults and accidents, the literature has focused on risk assessment and mitigation strategies to improve safety and reliability in port operations \cite{Wang2021AnAO, Cao2023ResearchIM}.

In addition to the aforementioned areas, recent literature has also emphasized the importance of sustainability and green practices in port operations \cite{Izaguirre2020ClimateCR, Lim2019PortSA}. With increasing global awareness of environmental issues, there has been a growing focus on developing sustainable and eco-friendly port operations. Sustainable port initiatives aim to minimize the environmental impact of port activities, reduce carbon emissions, and promote the use of renewable energy sources.

Despite the significant work on port operations, there is an absence of studies on models for predicting and classifying Total Time and Delay Time of vessels at ports, which highlights the need for research in this area \cite{h1, h3}. By addressing this gap, the proposed study aims to contribute insightful information for the decision-making processes in port management, ultimately improving the efficiency and effectiveness of port operations. This information will be crucial for solving other port related problems such as the BAP, as accurate predictions of Total Time and Delay Time can enhance the applicability of BAP solutions in real-world port environments. Moreover, these insights can support management decisions related to the planning, allocation, and scheduling of loading and unloading operations, contributing to overall operational excellence in port logistics.

Prediction and classification both play important roles in this context. Classification is a relatively easier task to complete and is useful for implementing measures that impact operations on a global scale. In contrast, prediction, though more challenging, is more valuable and precise, as it can help improve operations by addressing minute details. A similar analogy applies to Total Time and Delay Time, as they are approximate values and depend on different operations and the information available. Depending on the modeling task and the variables used, different features impact them to varying extents, providing valuable insights in the predictive analysis of port operations. The combination of these approaches ensures a comprehensive strategy, where classification enables broad decision-making and prediction drives detailed, data-driven improvements. By addressing these dimensions, the study offers a well-rounded approach to enhancing port operations.

\section{Case Study and Methodology}


The Case study of Brazil's ports is from \cite{brazildata}. Brazil's marine logistics and port operations are crucial components of its economy due to the country's extensive coastline and reliance on maritime transport for international trade. Brazil boasts one of the world's most extensive coastlines, stretching over 7,400 kilometers, with numerous ports facilitating the import and export of goods. The country's ports play a pivotal role in connecting Brazil to global markets, serving as crucial gateways for international trade \cite{Cabral2018Efficiency, p3}. However, port logistics in Brazil are complex due to inflexible regulations and the need for extensive information, impacting delivery times and costs \cite{b3, b5}. Overall, the efficient functioning of Brazil's marine logistics and port operations is essential for the country's economic prosperity, trade competitiveness, and integration into the global economy \cite{b1, b2}. It represents a critical component of Brazil's infrastructure and contributes significantly to its overall development.


We have translated the data from Portuguese to English. The data encompasses a wide array of characteristics related to maritime cargo transportation and port operations from a period of 11/2017-12/2018. They are categorized into four main groups: Cargo characteristics, Geographic characteristics, Operation characteristics, and Stay (Total) time characteristics. 

Starting with Cargo characteristics, the dataset includes details such as the type of cargo (ST Nature Cargo), the quantity of cargo moved in Twenty-foot Equivalent Units (TEU), the net weight of containerized cargo (VL Containerized Load Weight), and the overall weight of transported cargo in tons (Moved Value). These variables provide essential insights into the nature and volume of cargo being transported, which are crucial for understanding the logistical demands of port operations.

Moving on to Geographic characteristics, the dataset encompasses information about the source and destination of the cargo, the classification of inland transportation routes, the specific mooring berth utilized, the name of the mooring port and port complex, the city and province where the port is located, and the geographical region (such as Southeast, Northeast, North, or South). These variables offer geographic context and detail the locations and routes involved in the transportation process, which can impact logistical planning and operational decision-making.

The Operation characteristics section includes variables that detail various aspects of the cargo operations themselves. This includes the classification of cargo operations, the status of containers (whether they are full, empty, or carrying loose cargo), indicators for cabotage and long course operations, the sense of the operation (whether it is disembarked or embarked), the type of port authority (public or private), the classification of mooring navigation, and the nationality of the shipowner (whether Brazilian or foreign). These variables provide insights into the operational dynamics and regulatory aspects of cargo transportation.

Finally, the Stay (Total) time characteristics section includes variables related to the duration of a vessel's stay in port. This encompasses details such as the arrival date and hour of the ship, the mooring date and hour, the unblocking date and hour, the start date, end date and hours of operations, the month of mooring, and various time durations (T1-Wait time for the moat, T2-Wait time for start of operation, T3-Operation time, T4-Wait time for unberthing, T2+T3+T4-Mooring time, and T1+T2+T3+T4-Total Time) that reflect different aspects of the vessel's stay at port. These variables are critical for understanding the temporal aspects of port operations, including the time spent waiting, in operation, and overall stay duration. We define Delay Time as Total Time-Operation time (T1+T2+T4).

\begin{figure}[htbp]
	\centering
	\includegraphics[width=75mm]{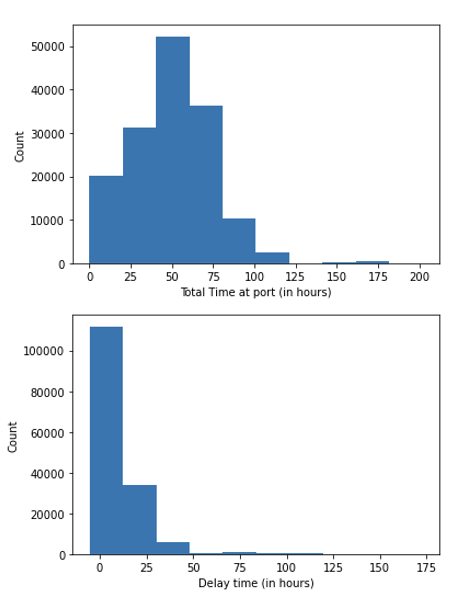} 
	\caption{Distribution of Total Time and Delay Time at port}\label{Fig:dt_tt}
\end{figure}

We define the Total Time (stay time) class using Fig. \ref{Fig:dt_tt}, similar to \cite{brazildata}. Class 0 (Low) is defined for vessels with Total Time at port for 0-25 hours, Class 1 (Medium) is defined for 25-50 hours, Class 2 (High) includes 50-75 hours and Class 3 (Very High) is $>$75 hours. For Delay Class, we use the plots as well as summary statistics (mean Delay Time$\approx$ 10 hours, variance of Delay Time$\approx$125), and define class 0 as Delay Time $<$ 24 hours and Class 1 as Delay Time $>$ 24 hours.

In the realm of supervised learning for prediction and classification, tree-based methods, statistics-based methods, and deep learning represent three prominent approaches, each with its own strengths and applications. Tree-based algorithms, such as Random Forest, Extreme Gradient Boosting (XGBoost), AdaBoost, Light Gradient Boosting Machine (LightGBM), and Extra Trees, are known for their ability to handle complex relationships in data and provide robust predictions. Random Forest, for instance, constructs multiple decision trees and aggregates their outputs to improve accuracy and reduce overfitting \cite{m1}. XGBoost is a gradient boosting algorithm that excels in optimizing complex objective functions and has been widely used in competitions and industry applications for its performance \cite{m2}. AdaBoost focuses on iteratively boosting the performance of weak learners to achieve strong predictive models \cite{m3}. LightGBM is known for its efficiency in handling large datasets and has become popular for its speed and accuracy \cite{m4}. Extra Trees or Extremely Randomized Trees is an ensemble learning method that builds multiple decision trees and averages their predictions, providing robustness against noise and overfitting \cite{m5}. Additionally, feature importance analysis in tree-based models helps identify the most influential variables in the prediction process, providing insights into the underlying patterns in the data \cite{m6}.

On the other hand, statistics-based methods like regression, Support Vector Machines (SVM), Bayesian classifiers, k-Nearest Neighbors (kNN), and Elastic Net offer a different set of tools for prediction and classification tasks. Regression techniques, including linear regression and its variants, are widely used for predicting continuous outcomes based on the relationships between variables \cite{m7}. SVM is a powerful predictive method that aims to find the optimal hyperplane that best separates data points into different regions in a high-dimensional space \cite{m8}. Bayesian classifiers are based on Bayes' theorem and are probabilistic models that calculate the probability of a sample belonging to a particular class \cite{m9}. kNN is a simple yet effective algorithm that classifies a data point based on the majority class of its nearest neighbors in the feature space \cite{m10}. Elastic Net is a regularization technique combining L1 (Lasso) and L2 (Ridge) penalties, which is particularly useful for feature selection and dealing with multicollinearity in regression and classification problems \cite{m11}. Each of these statistics-based methods has its own assumptions, strengths, and limitations, making them suitable for different types of datasets and prediction tasks.

In contrast, deep learning approaches like neural networks \cite{d1, d3} and ResNet (Residual Network) \cite{d2} can learn complex relationships present in the data. Neural networks consist of layers of interconnected nodes that can model intricate patterns and interactions between variables, making them highly effective for both classification and prediction tasks. ResNet is a type of deep neural network that introduces skip connections to enable the training of very deep models by mitigating the vanishing gradient problem, allowing it to capture even more complex data relationships. These deep learning models are particularly powerful for handling large and high-dimensional datasets, offering a complementary approach to traditional statistics-based methods.

In both prediction and classification tasks, various metrics are used to evaluate the performance of machine learning models. For the prediction tasks, we use common metrics like Root Mean Squared Error (RMSE), Mean Absolute Error (MAE), Mean Squared Error (MSE), R-squared ($R^2$), Root Mean Squared Logarithmic Error (RMSLE), and Mean Absolute Percentage Error (MAPE). MAE measures the average magnitude of the errors between predicted and actual values. MSE is the average of the squared differences between predicted and actual values (RMSE is the square root of MSE). $R^2$ represents the proportion of the variance in the dependent variable that is predictable from the independent variables. RMSLE is particularly useful for tasks where the target variable spans several orders of magnitude. MAPE measures the percentage difference between predicted and actual values. For all the metrics, lower values indicate better fit whereas for $R^2$ values close to 1 are considered good.

For classification tasks, we use metrics such as Accuracy, Precision, Recall, Area Under the Receiver Operating Characteristic Curve (AUC), F1 score, and Cohen's Kappa. All these values can be derived from the confusion matrix. Accuracy measures the proportion of correctly classified instances over the total number of instances. Precision measures the proportion of true positive predictions among all positive predictions, focusing on the accuracy of positive predictions. Recall (also known as sensitivity) measures the proportion of true positive predictions among all actual positives, focusing on the model's ability to detect positive instances. AUC measures the model's ability to distinguish between the classes. F1 score is the harmonic mean of precision and recall, providing a balance between the two metrics. Cohen's Kappa measures the agreement between predicted and actual classifications, accounting for the possibility of the agreement occurring by chance. For all the metrics, higher values indicate good performance.

We use Cross-validation \cite{Kohavi1995ASO} to assess the model's performance and tune its hyperparameters by repeatedly splitting the dataset into subsets for training and validation. It helps prevent overfitting by evaluating the model's performance on multiple subsets of the data, allowing for more reliable hyperparameter tuning and regularization.

Feature analysis is performed using Feature Importance \cite{fi} and SHAP \cite{shap}. They provide valuable insights into how individual features impact the model and the outcomes.  Feature Importance ranks the variables based on their contribution to the model's performance, helping to identify which factors are most influential in the decision-making process. SHAP, on the other hand, explains how each feature impacts the outcome for each individual instance, highlighting both the global and the local effects. However, SHAP does not measure the contribution of a feature to the overall model performance but instead focuses on the direction and magnitude of a feature's impact on specific outcomes. This dual approach enables a deeper understanding of the underlying dynamics in the data, allowing for more informed decisions and targeted improvements in complex systems like port operations.

In the context of predictive analytics for port operations, the aforementioned machine learning and deep learning methods are applied to predict and classify Total Time and Delay Time, which are critical factors in port logistics. By leveraging these techniques, we gain valuable insights into the intricate dynamics of port operations.

\section{Results}

We split the data (n=153331) into 80\% (122664) for training and 20\% (30667) for testing. We consider 24 (out of 35) features as independent variables and the other features related to calculation of Total Time and Delay Time at the port are removed. 17 of them are categorical variables. These categorical variables are transformed using one hot encoding (low cardinality variables like Month, Type of Mooring, Geographic Region, etc) or target encoding (high cardinality variables like Berth, Source, Destination, Port, etc). To ensure we don't overfit and to pick the best hyperparameters, we perform 10-fold cross-validation (CV). We first look into the prediction of Total Time and Delay Time, followed by the classification of Total Time and Delay Time. The experiments were performed on an 8-core M2 CPU.

Also, we perform feature analysis for a comprehensive understanding of the key factors influencing port operations. Addressing these factors is crucial for optimizing the Berth Allocation Problem (BAP), Predictive Model Scheduling (PMS), port performance management, and risk assessment. By enhancing these aspects, ports can achieve greater sustainability, aligning with the goal of promoting green practices.

\begin{table*}[htbp]
\centering
\begin{tabular}{|c|c|c|c|c|c|c|c|c|}
\hline Model & MAE & MSE & RMSE & $R^2$ & RMSLE & MAPE  \\
\hline \textbf{Random Forest} & 5.3336 & 108.9832 & \textbf{10.4377} & 0.8241 & 0.2992 & 0.1358 \\
\hline Extra Trees & 5.0287 & 121.0471 & 10.9998 & 0.8046 & 0.3112 & 0.1282  \\
\hline Extreme Gradient Boosting & 7.4763 & 125.7988 & 11.2152 & 0.7969 & 0.3343 & 0.1974 \\
\hline ResNet & 7.4643 & 126.6790 & 11.2551 & 0.7955 & 0.3418 & 0.2033 \\
\hline Neural Network & 7.551 & 127.9665 & 11.3122 & 0.7934 & 0.3547 & 0.2192 \\
\hline Light Gradient Boosting Machine & 8.3872 & 147.0024 & 12.1234 & 0.7627 & 0.3648 & 0.2256 \\
\hline Decision Tree & 5.4081 & 165.8789 & 12.8762 & 0.7322 & 0.3469 & 0.1378  \\
\hline K Nearest Neighbors & 8.1328 & 184.7665 & 13.5920 & 0.7017 & 0.3968 & 0.2074  \\
\hline Gradient Boosting & 10.4030 & 211.0671 & 14.5265 & 0.6593 & 0.4238 & 0.2728  \\
\hline Bayesian Ridge & 11.8099 & 273.8831 & 16.5479 & 0.5579 & 0.4772 & 0.3203  \\
\hline Ridge Regression & 11.8092 & 273.8808 & 16.5479 & 0.5579 & 0.4773 & 0.3202  \\
\hline Linear Regression & 11.8089 & 273.8825 & 16.5479 & 0.5579 & 0.4774 & 0.3201  \\
\hline Elastic Net & 12.8900 & 314.4793 & 17.7318 & 0.4924 & 0.5022 & 0.3847  \\
\hline Lasso Regression & 12.9111 & 315.2336 & 17.7531 & 0.4911 & 0.5023 & 0.3845  \\
\hline Huber Regressor & 12.5049 & 315.4952 & 17.7600 & 0.4907 & 0.5316 & 0.3412  \\
\hline AdaBoost & 18.0348 & 502.1311 & 22.3968 & 0.1889 & 0.6748 & 0.7616  \\
\hline
\end{tabular}
\caption{Comparing 10-fold CV results of different methods with multiple metrics (sorted using RMSE) for Total Time prediction at port}
\label{ttp_t}
\end{table*}

\begin{table*}[htbp]
\centering
\begin{tabular}{|c|c|c|c|c|c|c|c|c|} 
\hline Model & MAE & MSE & RMSE & $R^2$ & RMSLE & MAPE  \\
\hline \textbf{Random Forest} & 0.8102 & 17.8800 & \textbf{4.2179} & 0.6160 & 0.4997 & 0.5579  \\
\hline ResNet & 1.2444 & 17.8963 & 4.2304 & 0.6165 & 0.5610 & 0.6188 \\
\hline Extreme Gradient Boosting & 1.2285 & 18.1063 & 4.2470 & 0.6112 & 0.6103 & 0.7758 \\
\hline Extra Trees & 0.7304 & 19.8816 & 4.4488 & 0.5729 & 0.4981 & 0.5186  \\
\hline Light Gradient Boosting Machine & 1.3226 & 20.9450 & 4.5700 & 0.5502 & 0.6241 & 0.8042  \\
\hline Neural Network & 1.4564 & 24.6622 & 4.9661 & 0.4715 & 0.7127 & 0.8734  \\
\hline Decision Tree & 0.7571 & 26.6984 & 5.1538 & 0.4255 & 0.5464 & 0.5569  \\
\hline Gradient Boosting & 1.7824 & 29.9394 & 5.4656 & 0.3574 & 0.7811 & 0.9428 \\
\hline K Nearest Neighbors & 1.3783 & 35.7453 & 5.9735 & 0.2310 & 0.6818 & 0.7387  \\
\hline Bayesian Ridge & 2.3272 & 42.8552 & 6.5377 & 0.0815 & 0.9559 & 1.1375 \\
\hline Ridge Regression & 2.3273 & 42.8554 & 6.5378 & 0.0815 & 0.9564 & 1.1423  \\
\hline Linear Regression & 2.3271 & 42.8556 & 6.5378 & 0.0815 & 0.9564 & 1.1425  \\
\hline Elastic Net & 2.0899 & 44.3601 & 6.6513 & 0.0494 & 0.8636 & 0.9716  \\
\hline Lasso Regression & 2.1192 & 44.7360 & 6.6794 & 0.0414 & 0.8685 & 0.9946 \\
\hline Huber Regression & 1.2250 & 48.0613 & 6.9232 & -0.0299 & 0.7032 & 0.9955  \\
\hline AdaBoost & 3.2974 & 54.0987 & 7.3534 & -0.1708 & 1.1535 & 1.6455 \\
\hline
\end{tabular}
\caption{Comparing 10-fold CV results of different methods with multiple metrics (sorted using RMSE) for Delay Time prediction at port}
\label{dtp_t}
\end{table*}

\begin{figure}
\begin{subfigure}{\textwidth}
  \centering
  \includegraphics[width=85mm]{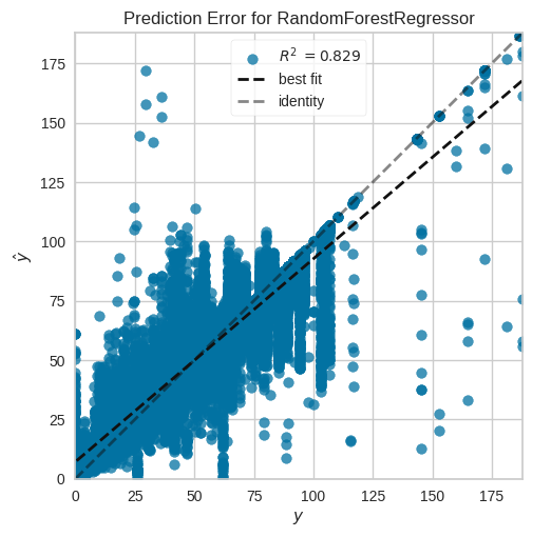}
  \caption{R-squared plot on test data for Total Time prediction}
  \label{fig:ttp_qq}
\end{subfigure}
\begin{subfigure}{\textwidth}
  \centering
  \includegraphics[width=95mm]{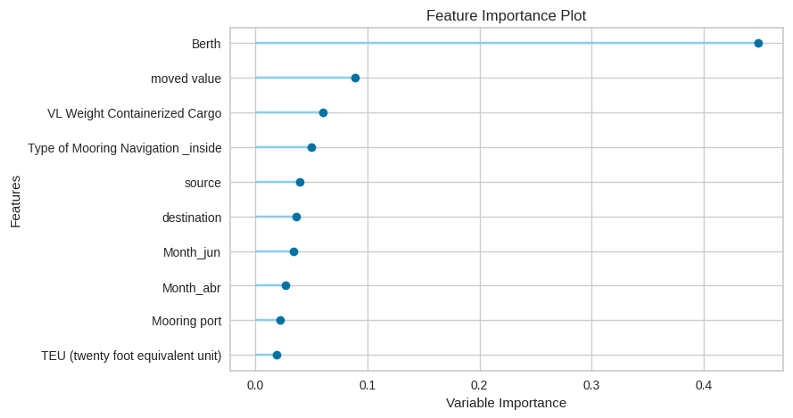}
  \caption{Random Forest Feature importance for of Total Time prediction}
  \label{fig:ttp_feat}
\end{subfigure}
\caption{Random Forest inference plots for Total Time prediction}
\label{fig:ttp_feat1111}
\end{figure}

Table \ref{ttp_t} shows the performance of different Machine Learning and Deep Learning approaches for Total Time prediction. We see a general trend that tree-based methods are performing better than statistical approaches on this task. This might be because of the complex nature of the data and their superiority in handling categorical variables. Surprisingly, Neural network and ResNet performance is a bit underwhelming, the main reason for this is the presence of many categorical features in the data \cite{9512057, Grinsztajn2022WhyDT}. The High Cardinality, Non-Ordinal Nature, and Insufficient Sample Per Category are hampering the deep learning performance. We observe that Random Forest (RF) performs the best, has the lowest RMSE (10.4377), and also has better values in other metrics in comparison to the rest of the methods. Performance on the test set for RF is similar, where the RMSE is 10.2577 (other metrics are also in close range). This is visible from Fig. \ref{fig:ttp_qq}, where the predicted values and true values are near to the identity line with an $R^2$ value of 0.8287.

\begin{figure}
\begin{subfigure}{\textwidth}
  \centering
  \includegraphics[width=95mm]{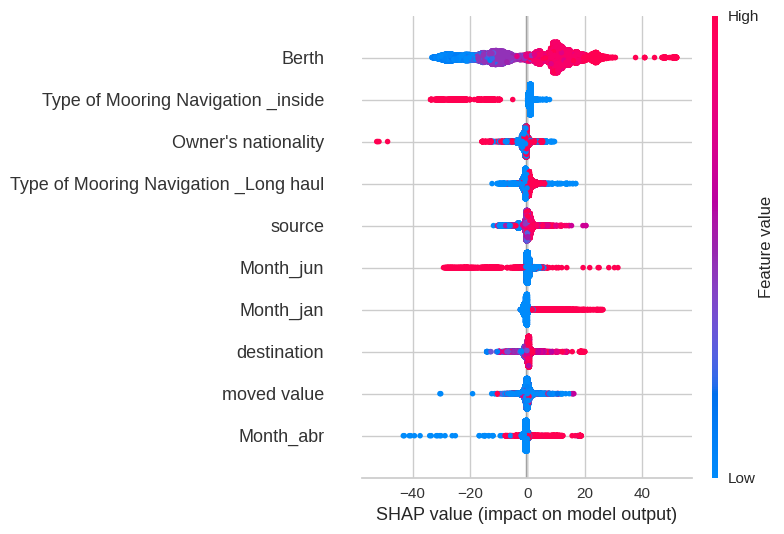}
  \caption{SHAP feature analysis plot for Total Time Prediction}
  \label{fig:ttp_qq1}
\end{subfigure}
\begin{subfigure}{\textwidth}
  \centering
  \includegraphics[width=95mm]{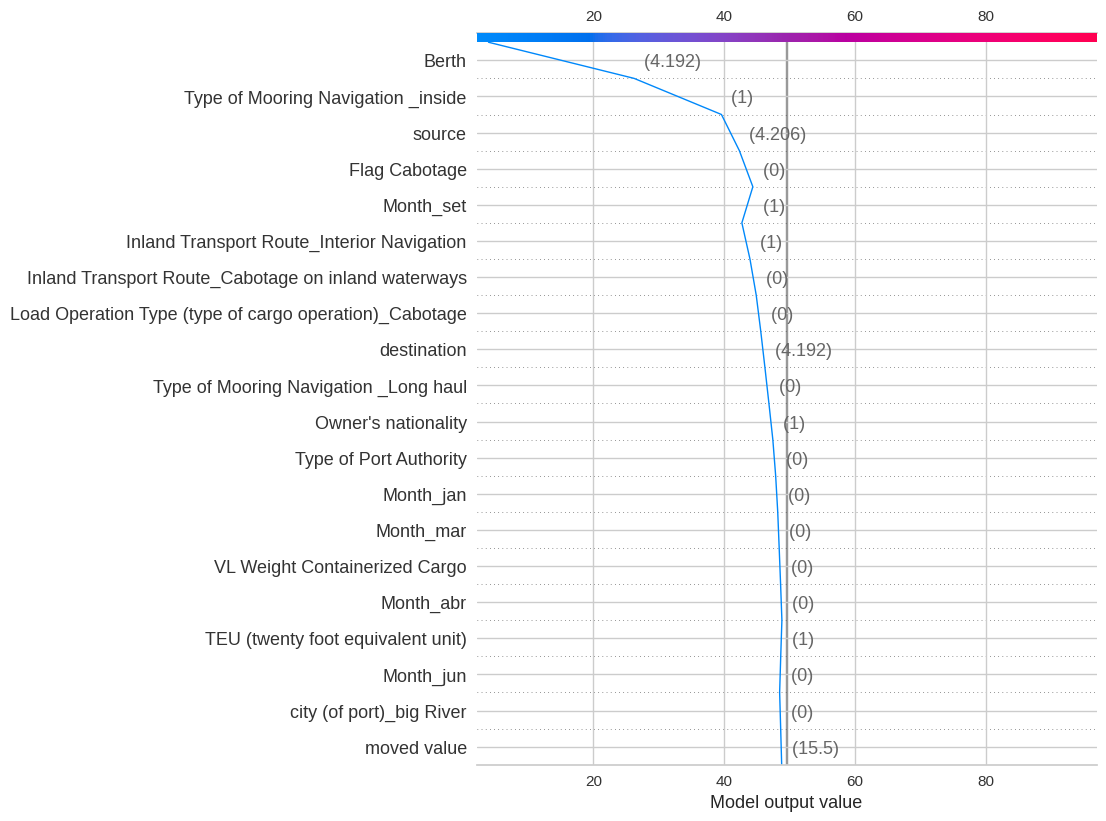}
  \caption{SHAP Decision plot for Total Time Prediction of a sample observation}
  \label{fig:ttp_qq2}
\end{subfigure}
\begin{subfigure}{\textwidth}
  \centering
  \includegraphics[width=95mm]{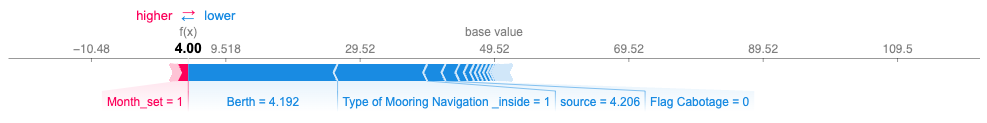}
  \caption{SHAP force plot for Total Time Prediction of a sample observation}
  \label{fig:ttp_qq3}
\end{subfigure}
\caption{SHAP Analysis for Total Time Prediction}
\label{fig:ttp_qqall}
\end{figure}

Fig. \ref{fig:ttp_feat} shows the feature importance for the Total Time prediction, and we can observe that Berth feature plays the most crucial role due to some berths being more busy and less efficient compared to others, followed by other features like Moved Value, VL Containerized Load Weight, Type of Mooring Navigation, Source, Destination, Month, etc. Since, Berths play a vital role in port operations, efficient berth management is essential for optimizing port capacity, throughput, and overall smooth operations. The SHAP analysis can be seen in Fig. \ref{fig:ttp_qqall}, where Fig. \ref{fig:ttp_qq1} indicates that features like Berth, Type of Mooring Navigation, and Nationality have a high impact on the outcomes for the random forest model. The horizontal spread indicates the magnitude of the effect that a feature has on the outcome whereas the color represents if the feature value is high or low. For example, the Berth feature (target encoded) denotes that, generally, a high value of this feature leads to a higher positive impact on the outcome. This means that it will lead to a higher Total Time value. Fig. \ref{fig:ttp_qq2} and Fig. \ref{fig:ttp_qq3} show the Decision and Force plot for a particular sample with true Total Time of 4.50 hours. The base value (49.52) represents the expected value of the model's output before any features are considered. The model output is 4, which is near the truth. We see features like Berth, Type of Mooring Navigation, and Source have the highest impact on getting the right output for this sample. Since Berth and Source are fixed features, changing the Type of Mooring Navigation for this sample might help to reduce the Total Time. Note that for a different sample, some other features might be important.

\begin{figure}
\begin{subfigure}{\textwidth}
  \centering
  \includegraphics[width=85mm]{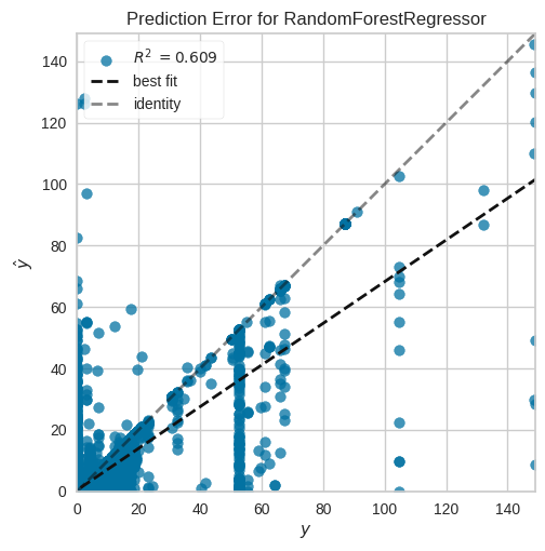}
  \caption{R-squared plot on test data for Delay Time prediction}
  \label{fig:dtp_qq}
\end{subfigure}
\begin{subfigure}{\textwidth}
  \centering
  \includegraphics[width=95mm]{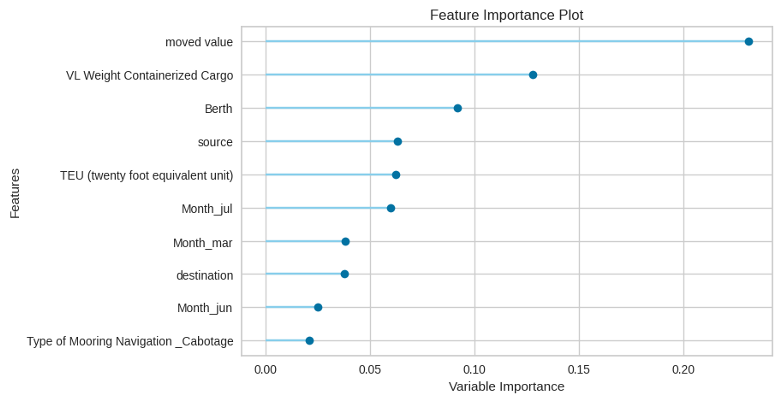}
  \caption{Random Forest Feature importance for of Delay Time prediction}
  \label{fig:dtp_feat}
\end{subfigure}
\caption{Random Forest inference plots for Delay Time prediction}
\label{fig:dtp_qq1111}
\end{figure}

We see a similar story from Table \ref{dtp_t} where the tree-based methods are performing better for predicting the Delay Time at the port. Random Forest (RF) again performs the best with an RMSE of 4.2179. The test set performance for RF gives an RMSE of 4.3707 (other metrics are also in a similar range). From Fig. \ref{fig:dtp_qq}, we can see that the predicted and true values are not close to the identity line ($R^2$ is 0.6086). The prediction of Delay Time might be limited by the feature set, as it is known that weather information and other external factors play a key role with respect to delay.

\begin{figure}
\begin{subfigure}{\textwidth}
  \centering
  \includegraphics[width=95mm]{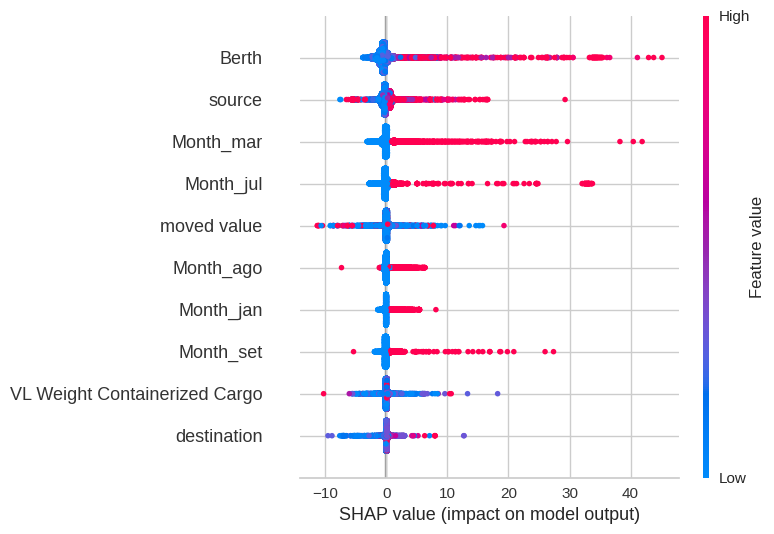}
  \caption{SHAP feature analysis plot for Delay Time Prediction}
  \label{fig:dtp_qq1}
\end{subfigure}
\begin{subfigure}{\textwidth}
  \centering
  \includegraphics[width=95mm]{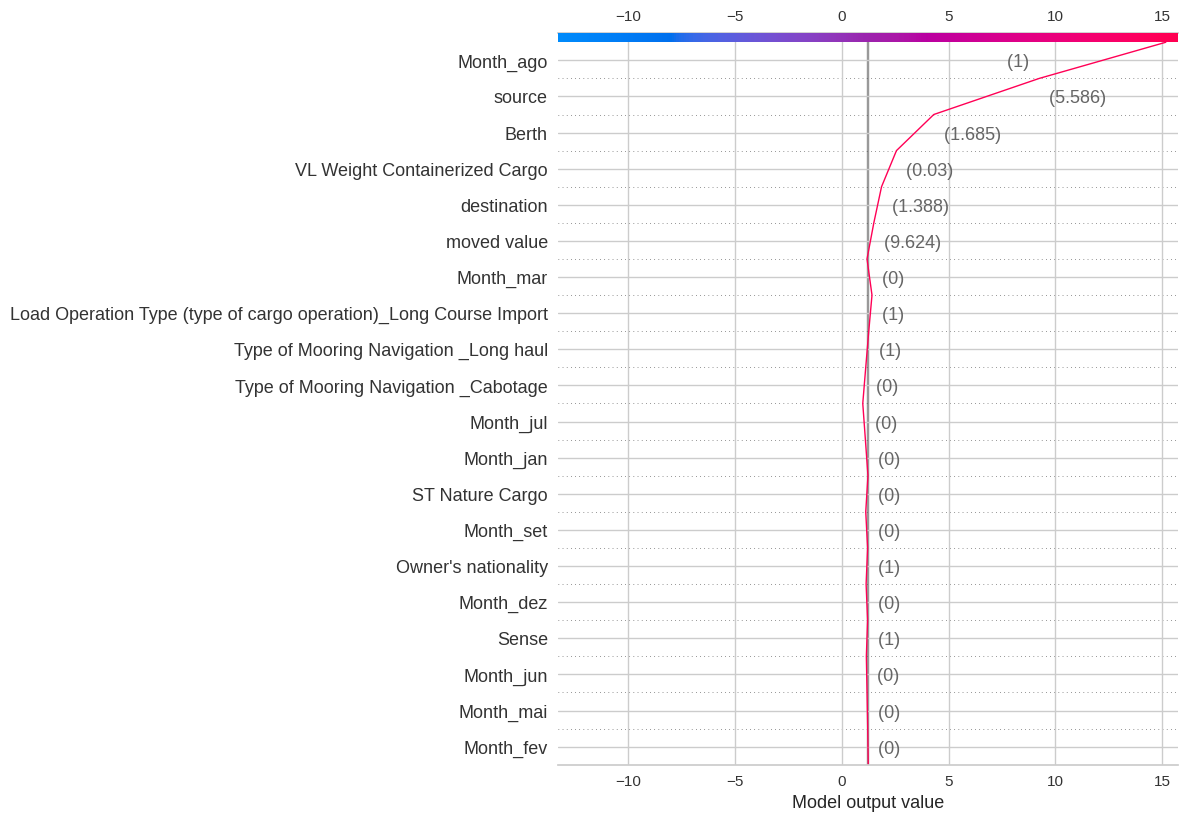}
  \caption{SHAP Decision plot for Delay Time Prediction of a sample observation}
  \label{fig:dtp_qq2}
\end{subfigure}
\begin{subfigure}{\textwidth}
  \centering
  \includegraphics[width=95mm]{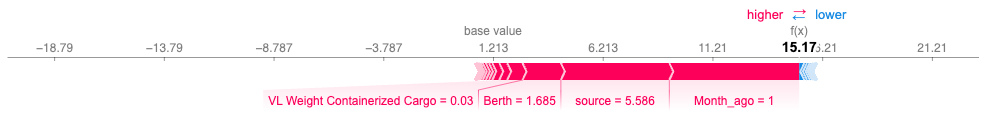}
  \caption{SHAP force plot for Delay Time Prediction of a sample observation}
  \label{fig:dtp_qq3}
\end{subfigure}
\caption{SHAP Analysis for Delay Time Prediction}
\label{fig:dtp_qqall}
\end{figure}

Fig. \ref{fig:dtp_feat} shows the feature importance for the Delay Time prediction, and we can observe that Moved Value, VL Weight Containerized Cargo, and Berth are key features followed by other features like Source, Destination, Month, etc. It is intuitive that vessels with higher Moved Value and VL Containerized Load Weight go through longer operations leading to build up in Delay Time. We can see the SHAP analysis in Fig. \ref{fig:dtp_qqall}, where Fig. \ref{fig:dtp_qq1} shows that the Berth, Source, Month, and Moved Value have the most impact on the outcomes for the random forest model. Fig. \ref{fig:dtp_qq2} and Fig. \ref{fig:dtp_qq3} show the Decision and Force plot for a particular sample with true Delay Time of 16 hours. We get a predicted value of 15.17, which is very close to the truth, and can see that features like Month, Source, and Berth have the most impact towards getting the correct output for this sample. Unfortunately, for this sample, there is no potential for the outcome to be changed as the top features are fixed.

\begin{table*}[htbp]
\centering
\begin{tabular}{|c|c|c|c|c|c|c|c|c|c|}
\hline Model & Accuracy & AUC & Recall & Precision & F1 & Kappa  \\
\hline \textbf{Extreme Gradient Boosting} & \textbf{0.8424} & 0.9716 & 0.8424 & 0.8426 & 0.8424 & 0.7797 \\
\hline Extra Trees & 0.8386 & 0.9453 & 0.8386 & 0.8386 & 0.8386 & 0.7743  \\
\hline Random Forest & 0.8360 & 0.9634 & 0.8360 & 0.8361 & 0.8360 & 0.7707  \\
\hline Decision Tree & 0.8220 & 0.8775 & 0.8220 & 0.8219 & 0.8219 & 0.7511  \\
\hline ResNet & 0.8165 & 0.8503 & 0.8165 & 0.8168 & 0.8166 & 0.7785  \\
\hline Neural Network & 0.8108 & 0.8369 & 0.8108 & 0.8111 & 0.8109 & 0.7628  \\
\hline Light Gradient Boosting Machine & 0.7978 & 0.9518 & 0.7978 & 0.7981 & 0.7977 & 0.7165 \\
\hline Gradient Boosting & 0.7210 & 0.9051 & 0.7210 & 0.7235 & 0.7190 & 0.6046  \\
\hline K Nearest Neighbors & 0.7160 & 0.8962 & 0.7160 & 0.7171 & 0.7156 & 0.6009  \\
\hline Linear Discriminant Analysis & 0.6344 & 0.8353 & 0.6344 & 0.6355 & 0.6281 & 0.4788 \\
\hline Ridge Classifier & 0.6296 & 0.8194 & 0.6296 & 0.6355 & 0.6117 & 0.4661  \\
\hline Logistic Regression & 0.5997 & 0.8135 & 0.5997 & 0.6183 & 0.5626 & 0.4212  \\
\hline Ada Boost & 0.5511 & 0.7245 & 0.5511 & 0.5447 & 0.5457 & 0.3687 \\
\hline SVM & 0.5412 & 0.7749 & 0.5412 & 0.5808 & 0.5159 & 0.3382 \\
\hline Naive Bayes & 0.5342 & 0.7675 & 0.5342 & 0.5769 & 0.5136 & 0.3538  \\
\hline Quadratic Discriminant Analysis & 0.3429 & 0.7174 & 0.3429 & 0.4488 & 0.2962 & 0.1611  \\
\hline
\end{tabular}
\caption{Comparing 10-fold CV results of different methods with multiple metrics (sorted using Accuracy) for Total Time classification at port}
\label{ttc_t}
\end{table*}

\begin{table*}[htbp]
\centering
\begin{tabular}{|c|c|c|c|c|c|c|c|c|c|}
\hline Model & Accuracy & AUC & Recall & Precision & F1 & Kappa \\
\hline \textbf{Extreme Gradient Boosting} & \textbf{0.9511} & 0.9798 & 0.6923 & 0.7507 & 0.7202 & 0.6935   \\
\hline Extra Trees & 0.9498 & 0.9365 & 0.7014 & 0.7348 & 0.7176 & 0.6901  \\
\hline Random Forest & 0.9491 & 0.9701 & 0.6853 & 0.7364 & 0.7098 & 0.6820  \\
\hline ResNet & 0.9464 & 0.7975 & 0.6157 & 0.7484 & 0.6755 & 0.6489 \\
\hline Decision Tree & 0.9459 & 0.8394 & 0.6907 & 0.7079 & 0.6991 & 0.6694 \\
\hline Neural Network & 0.9460 & 0.7827 & 0.5832 & 0.7674 & 0.6627 & 0.6208 \\
\hline Light Gradient Boosting Machine & 0.9389 & 0.9688 & 0.4375 & 0.7996 & 0.5652 & 0.5354   \\
\hline K Nearest Neighbors & 0.9228 & 0.8741 & 0.4426 & 0.6027 & 0.5104 & 0.4695   \\
\hline Gradient Boosting & 0.9166 & 0.8990 & 0.1295 & 0.7371 & 0.2196 & 0.1979  \\
\hline Ridge Classifier & 0.9097 & 0.8418 & 0.0360 & 0.5570 & 0.0676 & 0.0573   \\
\hline Logistic Regression & 0.9079 & 0.8080 & 0.0436 & 0.4363 & 0.0792 & 0.0638   \\
\hline Linear Discriminant Analysis & 0.9059 & 0.8159 & 0.1250 & 0.4385 & 0.1945 & 0.1607   \\
\hline Ada Boost & 0.9056 & 0.8277 & 0.0684 & 0.3893 & 0.1162 & 0.0917   \\
\hline SVM & 0.8135 & 0.8375 & 0.2273 & 0.3695 & 0.1117 & 0.0688  \\
\hline Naive Bayes & 0.5680 & 0.7600 & 0.8872 & 0.1606 & 0.2720 & 0.1394  \\
\hline Quadratic Discriminant Analysis & 0.1167 & 0.5137 & 0.9978 & 0.0932 & 0.1705 & 0.0049 \\
\hline
\end{tabular}
\caption{Comparing 10-fold CV results of different methods with multiple metrics (sorted using Accuracy) for Delay Time classification at port}
\label{dtc_t}
\end{table*}

\begin{figure}
\begin{subfigure}{\textwidth}
  \centering
  \includegraphics[width=85mm]{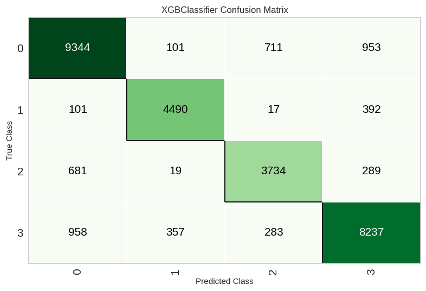}
  \caption{Confusion matrix on test data for Total Time classification}
  \label{fig:ttc_cm}
\end{subfigure}
\begin{subfigure}{\textwidth}
  \centering
  \includegraphics[width=95mm]{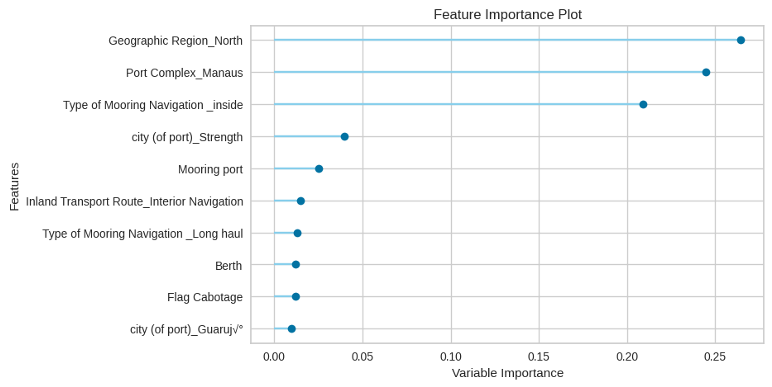}
  \caption{Random Forest Feature importance for of Total Time classification}
  \label{fig:ttc_feat}
\end{subfigure}
\caption{Random Forest inference plots for Total Time classification}
\label{fig:ttc_cm1111}
\end{figure}

In the classification task, the performance of various Machine Learning methods for (Multi-Class) classification of Total Time is depicted in Table \ref{ttc_t}. A consistent trend is observed where tree-based methods outperform statistical approaches as well as deep learning methods again, potentially due to the complex nature of the data and high number of categorical features. Specifically, Extreme Gradient Boosting (XGBoost) demonstrates the best performance, achieving the highest Accuracy (0.8424) and superior values across other metrics compared to alternative methods. This is 10\% higher than the result in \cite{brazildata}. The test set results further confirm XGBoost's efficacy with an Accuracy of 0.8415 and other metrics exhibiting similarly favorable performance, derived from the confusion matrix in Fig. \ref{fig:ttc_cm}. Also, we observe that is it easier to detect the outer classes, Class 0 (Low) and Class 3 (Very High), compared to the other two.

\begin{figure}
\begin{subfigure}{\textwidth}
  \centering
  \includegraphics[width=95mm]{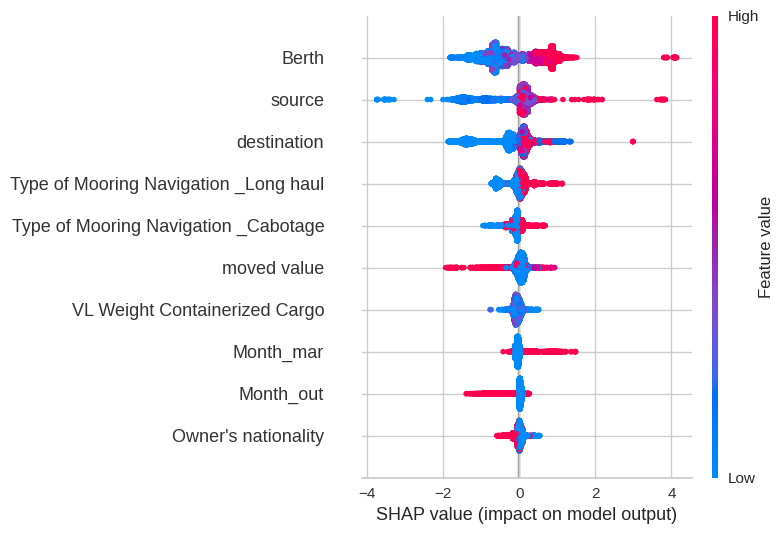}
  \caption{SHAP feature analysis plot for  Total Time classification}
  \label{fig:dtc_cm11}
\end{subfigure}
\begin{subfigure}{\textwidth}
  \centering
  \includegraphics[width=95mm]{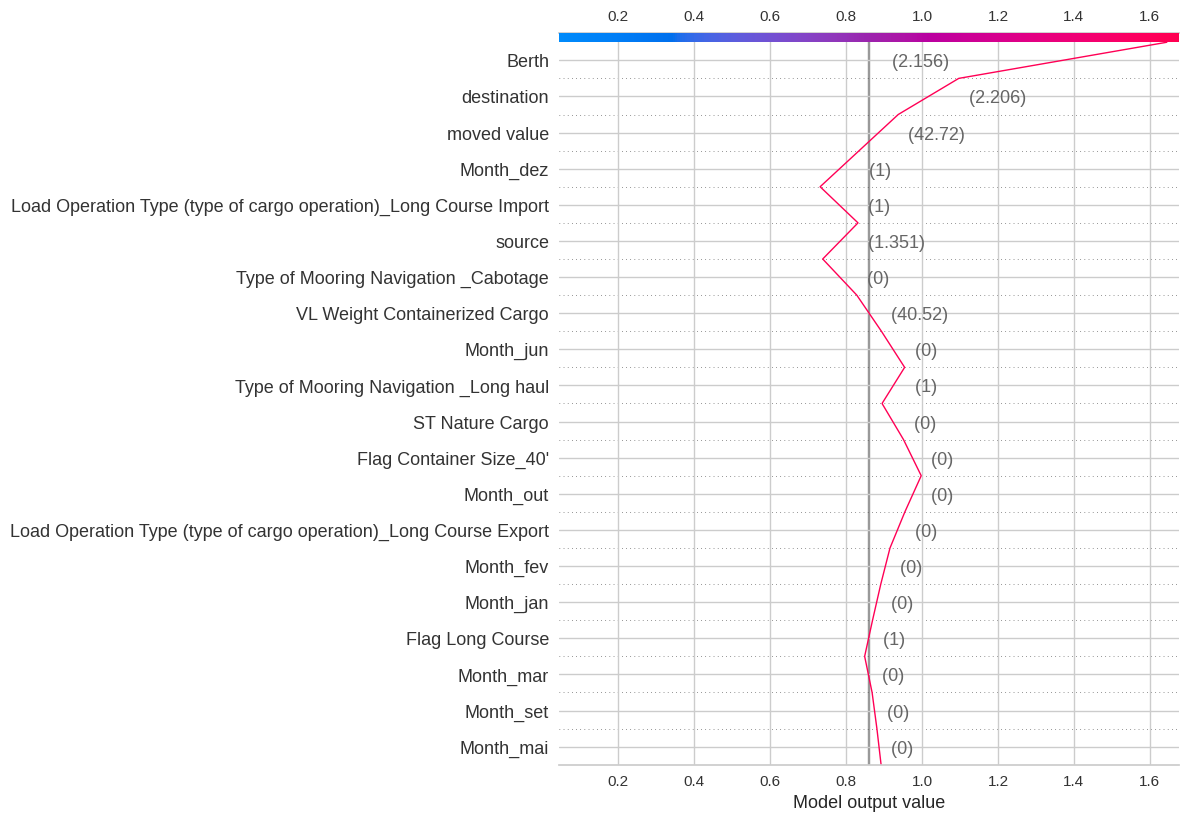}
  \caption{SHAP Decision plot for  Total Time classification of a sample observation}
  \label{fig:dtc_cm21}
\end{subfigure}
\begin{subfigure}{\textwidth}
  \centering
  \includegraphics[width=95mm]{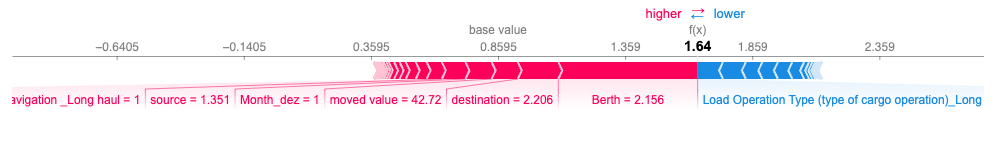}
  \caption{SHAP force plot of a sample observation for Class 1}
  \label{fig:dtc_cm22}
\end{subfigure}
\caption{SHAP Analysis for Total Time classification}
\label{fig:dtc_cmall1}
\end{figure}

Additionally, Fig. \ref{fig:ttc_feat} showcases the key features influencing Total Time classification. Notably, the Geographical Region (North), Port Complex (Manaus), Type of Mooring Navigation (inside) emerge as the most influential features, followed by factors such as City, Mooring port, Inland Transportation route, Type of Mooring Navigation (Long Haul), Berth, etc. The significance of the geographical features is related to the huge coastline of Brazil and the complex dynamics of port operations. The SHAP analysis is illustrated in Fig. \ref{fig:dtc_cmall1}. Specifically, Fig. \ref{fig:dtc_cm11} demonstrates that features such as Berth, Destination, Moved Value, and Month have the most significant impact on the outcomes of the XGBoost model. Each sample will have 4 Decision and Force plots, one for each class, since the sample considered has a class of 1, we shared the plots of Class 1. Fig. \ref{fig:dtc_cm21} and Fig. \ref{fig:dtc_cm22} present the Decision and Force plots for Class 1 for a particular sample with a true Total Time class as 1 (Medium). The model output is 1.64, which is the highest among the four classes and hence this sample is correctly classified as class 1. These plots reveal that features like Berth, Destination, and Moved Value play a crucial role in producing the accurate classification for this specific sample. Reducing the Moved Value for this sample should push it towards Class 0 (Low), which intuitively makes sense but may not be feasible.



\begin{figure}
\begin{subfigure}{\textwidth}
  \centering
  \includegraphics[width=85mm]{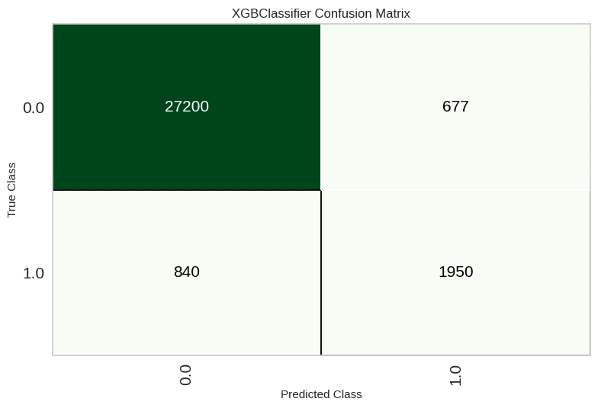}
  \caption{Confusion matrix on test data for Delay Time classification}
  \label{fig:dtc_cm}
\end{subfigure}
\begin{subfigure}{\textwidth}
  \centering
  \includegraphics[width=95mm]{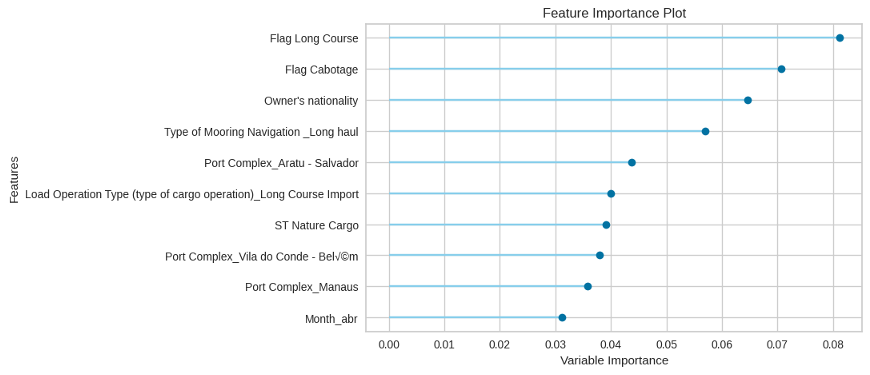}
  \caption{Random Forest Feature importance for of Delay Time classification}
  \label{fig:dtc_feat}
\end{subfigure}
\caption{Random Forest inference plots for Delay Time classification}
\label{fig:dtc_feat11111}
\end{figure}

Table \ref{dtc_t} presents the binary classification performance on Delay Time. The tree-based methods are again outperforming other approaches. XGBoost gives the best performance by achieving the highest Accuracy (0.9511) and best values across the different metrics compared to other methods. XGBoost achieves an Accuracy of 0.9505 on the test set and other metrics showing close performance, as illustrated by the confusion matrix in Fig. \ref{fig:dtc_cm}.

\begin{figure}
\begin{subfigure}{\textwidth}
  \centering
  \includegraphics[width=95mm]{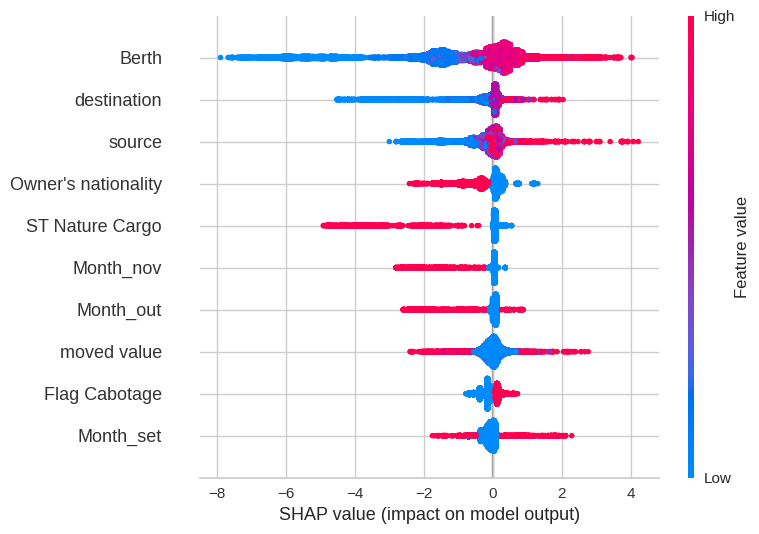}
  \caption{SHAP feature analysis plot for Delay Class}
  \label{fig:dtc_cm1}
\end{subfigure}
\begin{subfigure}{\textwidth}
  \centering
  \includegraphics[width=95mm]{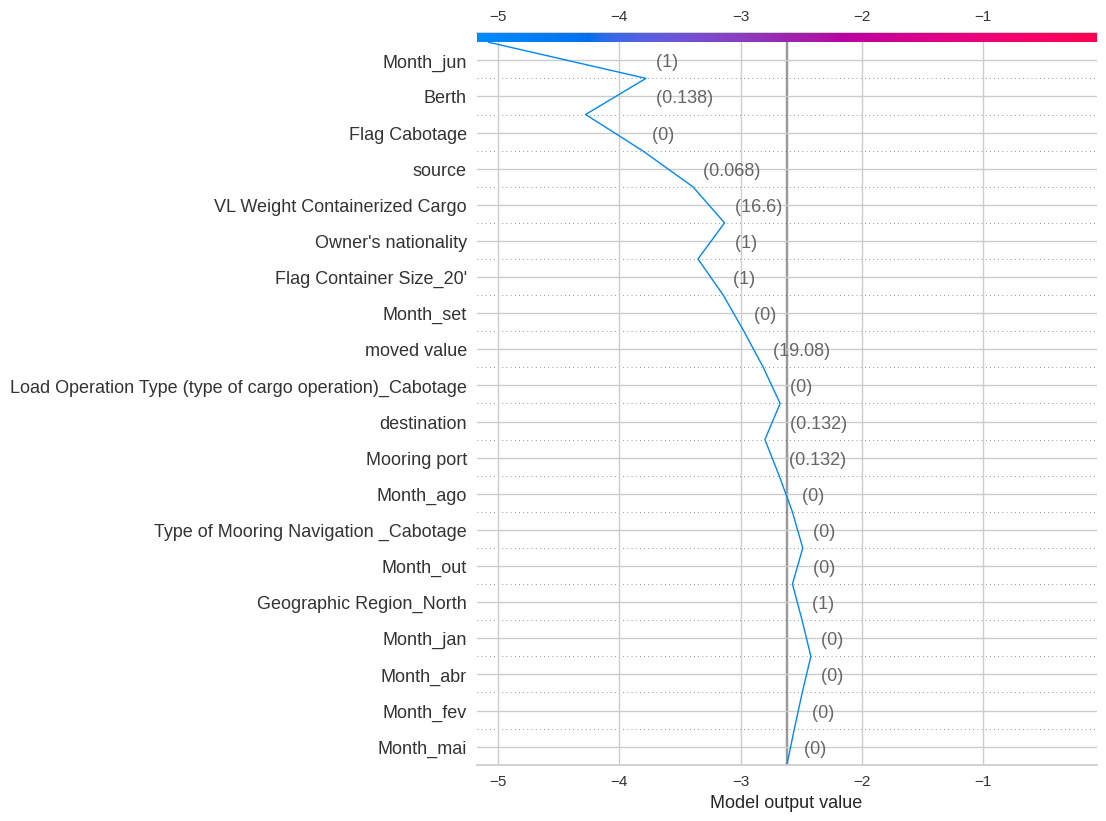}
  \caption{SHAP Decision plot for Delay Class of a sample observation}
  \label{fig:dtc_cm2}
\end{subfigure}
\begin{subfigure}{\textwidth}
  \centering
  \includegraphics[width=95mm]{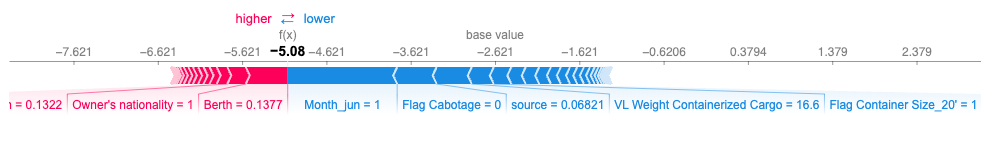}
  \caption{SHAP force plot for Delay Class of a sample observation}
  \label{fig:dtc_cm3}
\end{subfigure}
\caption{SHAP Analysis for Delay Class}
\label{fig:dtc_cmall}
\end{figure}

Furthermore, Fig. \ref{fig:dtc_feat} highlights the key features influencing Delay Time classification, factors such as Flag Long Course, Flag Cabotage, Owner's Nationality, and Type of Mooring Navigation (Long Haul) emerge as the most influential. This is followed by variables like Port Complex, Load Operation Type, ST Cargo Nature, and Month. Features related to operation characteristics are having the most impact on the Delay class. Fig. \ref{fig:dtc_cmall} shows the SHAP analysis for Delay Class, where Fig. \ref{fig:dtc_cm1} indicates that the most significant impact on the XGBoost output is by the features Berth, Source, destination, and Nationality. Figs. \ref{fig:dtc_cm2} and \ref{fig:dtc_cm3} show the Decision plot and Force plot for a specific sample, where the actual delay class is 0. The SHAP output is -5.08, indicating the class is class 0 since it has a low value, with features such as Month, Berth, and Flag Cabotage having the most influence on this sample. Since this sample is Class 0, meaning no delay, we don't want to change any feature in this case.






\section{Conclusion}


In conclusion, this development of a comprehensive prediction and classification solution for estimating vessel Stay (Total) Time and Delay Time in maritime logistics represents a significant step forward in addressing the complexities of port operations. Our study highlights the use and importance of predictive analysis for port operations using Brazilian port data. This can easily be extended to benefit port operations worldwide, offering opportunities for improved efficiency and effectiveness in maritime logistics on a global scale. Notably, tree-based methods exhibit the best performance, particularly in predicting Total Time and Delay Class. By providing tools to accurately predict and classify vessel behavior, this research contributes greatly to improving our understanding of various methods, which can be used to make more efficient and effective predictive analytics based decision-making in port management. The proposed solution has wide-ranging applications and integration into different aspects of maritime logistics, from ship scheduling to service delay predictions to risk assessment, demonstrating its potential to enhance overall port performance and optimize supply chain operations.

Additionally, the incorporation of feature analysis through feature importance enriches our understanding of the factors influencing maritime logistics, where predictions are influenced by Berth and Cargo Characteristics features while classification tasks are impacted by Geographical features (Total Time) and operation characteristic features (Delay Time). This offers insights that can be leveraged to further improve the efficiency of port operations and make them more resilient. Since Berth plays a key role in the prediction of Total Time and Delay Time, using a Top-Bottom approach for developing individual models for ports and then berths would be beneficial in understanding how these ports and berths can be made more efficient. SHAP (SHapley Additive exPlanations) analysis plays a crucial role in this process by helping us understand how these features contribute to the outcomes, providing deeper insights into each individual case. This allows for the identification of specific mitigation steps that can be taken accordingly. We observe that Berth, Source, and Destination, and Month are usually coming out as key features impacting outcomes. Learning from this, we can suggest an optimal Berth, Source, Destination combination and an ideal travel month if operationally feasible. SHAP helps us address the predictive analytics of port operations at both the global and local levels, ensuring that the models are both comprehensive and precise. Overall, this comprehensive approach, enhanced by Feature Importance and SHAP analysis, contributes to a more nuanced understanding of port logistics and supports the development of more effective and resilient port management strategies.

Deep learning methods do not currently perform well due to the abundance of categorical data. In the future, we plan to address this issue by exploring alternative methods for handling categorical data, such as binary encoding,  deep embedding (Cat2Vec), and other techniques, to see if they can improve the performance of deep learning models. We would like to focus on refining the prediction and classification models to account for additional variables and complexities in maritime logistics as a key area of future work. It could involve integrating real-time data streams, such as weather forecasts, vessel tracking, and other vessel information, to enhance the accuracy and responsiveness of the models. These expanded features will lead to enhanced feature analysis as well. Furthermore, exploring the application of more advanced techniques may offer new avenues for improving the modeling capabilities. Additionally, efforts to expand the scope of the research beyond individual ports to encompass broader regional or global maritime networks could provide valuable insights into optimizing port operations within larger logistical contexts. Overall, continued research in this area has the potential to drive significant advancements in maritime logistics, leading to more efficient and resilient port operations in the future.

\bibliographystyle{apalike}
\bibliography{ref}

\begin{thebibliography}{}

\bibitem[Abiodun et~al., 2018]{d1}
Abiodun, O.~I., Jantan, A.~B., Omolara, A.~E., Dada, K.~V., Mohamed, N., and Arshad, H. (2018).
\newblock State-of-the-art in artificial neural network applications: A survey.
\newblock {\em Heliyon}, 4.

\bibitem[Abreu et~al., 2022]{brazildata}
Abreu, L., Maciel, I., Alves, J., Braga, L., and Pontes, H. (2022).
\newblock A decision tree model for the prediction of the stay time of ships in brazilian ports.
\newblock {\em Engineering Applications of Artificial Intelligence}.

\bibitem[Altman, 1992]{m10}
Altman, N.~S. (1992).
\newblock An introduction to kernel and nearest-neighbor nonparametric regression.
\newblock {\em The American Statistician}, 46:175--185.

\bibitem[Bautista-S{\'a}nchez et~al., 2019]{BautistaSnchez2019StatisticalAI}
Bautista-S{\'a}nchez, R., Barbosa-Santill{\'a}n, L.~I., and S{\'a}nchez-Escobar, J.~J. (2019).
\newblock Statistical approach in data filtering for prediction vessel movements through time and estimation route using historical ais data.
\newblock In {\em Mexican International Conference on Artificial Intelligence}.

\bibitem[Bierwirth and Meisel, 2015]{Bierwirth2015AFS}
Bierwirth, C. and Meisel, F. (2015).
\newblock A follow-up survey of berth allocation and quay crane scheduling problems in container terminals.
\newblock {\em Eur. J. Oper. Res.}, 244:675--689.

\bibitem[Breiman, 2001]{m1}
Breiman, L. (2001).
\newblock Random forests.
\newblock {\em Machine Learning}, 45:5--32.

\bibitem[Cabral and Ramos, 2018]{Cabral2018Efficiency}
Cabral, A. M.~R. and Ramos, F.~S. (2018).
\newblock Efficiency container ports in brazil: A dea and fdh approach.
\newblock {\em The Central European Review of Economics and Management (CEREM)}, 2(1):43--64.

\bibitem[Cacho et~al., 2021]{b2}
Cacho, J., Tokarski, A., Thomas, E., and Chkoniya, V. (2021).
\newblock {\em Port Dada Integration: Opportunities for Optimization and Value Creation}, pages 1--22.
\newblock IGI Global.

\bibitem[Cao et~al., 2023]{Cao2023ResearchIM}
Cao, Y., Wang, X., Yang, Z., Wang, J., Wang, H., and Liu, Z. (2023).
\newblock Research in marine accidents: A bibliometric analysis, systematic review and future directions.
\newblock {\em Ocean Engineering}.

\bibitem[Chen and Guestrin, 2016]{m2}
Chen, T. and Guestrin, C. (2016).
\newblock Xgboost: A scalable tree boosting system.
\newblock {\em Proceedings of the 22nd ACM SIGKDD International Conference on Knowledge Discovery and Data Mining}.

\bibitem[Chu et~al., 2024]{h1}
Chu, Z., Yan, R., and Wang, S. (2024).
\newblock Vessel turnaround time prediction: A machine learning approach.
\newblock {\em Ocean \& Coastal Management}.

\bibitem[Cortes and Vapnik, 1995]{m8}
Cortes, C. and Vapnik, V.~N. (1995).
\newblock Support-vector networks.
\newblock {\em Machine Learning}, 20:273--297.

\bibitem[Costa et~al., 2022]{b1}
Costa, W., Soares-Filho, B., and Nobrega, R. (2022).
\newblock Can the brazilian national logistics plan induce port competitiveness by reshaping the port service areas?
\newblock {\em Sustainability}, 14:14567.

\bibitem[da~Veiga~Lima and de~Souza, 2022]{b3}
da~Veiga~Lima, F. and de~Souza, D. (2022).
\newblock Climate change, seaports, and coastal management in brazil: An overview of the policy framework.
\newblock {\em Regional Studies in Marine Science}, 52:102365.

\bibitem[Dahouda and Joe, 2021]{9512057}
Dahouda, M.~K. and Joe, I. (2021).
\newblock A deep-learned embedding technique for categorical features encoding.
\newblock {\em IEEE Access}, 9:114381--114391.

\bibitem[Dimitrios et~al., 2023]{p2}
Dimitrios, D., Nikitakos, N., Papachristos, D., and Dalaklis, A. (2023).
\newblock {\em Opportunities and Challenges in Relation to Big Data Analytics for the Shipping and Port Industries}, pages 267--290.
\newblock Palgrave Macmillan, Cham.

\bibitem[Dulebenets, 2018]{Dulebenets2018ACM}
Dulebenets, M.~A. (2018).
\newblock A comprehensive multi-objective optimization model for the vessel scheduling problem in liner shipping.
\newblock {\em International Journal of Production Economics}, 196:293--318.

\bibitem[Eriksen et~al., 2006]{Eriksen2006MaritimeTM}
Eriksen, T., Hoye, G., Narheim, B.~T., and Meland, B.~J. (2006).
\newblock Maritime traffic monitoring using a space-based ais receiver.
\newblock {\em Acta Astronautica}, 58:537--549.

\bibitem[Farag and {\"O}lçer, 2020]{Farag2020TheDO}
Farag, Y.~B. and {\"O}lçer, A.~I. (2020).
\newblock The development of a ship performance model in varying operating conditions based on ann and regression techniques.
\newblock {\em Ocean Engineering}, 198:106972.

\bibitem[Freund and Schapire, 1997]{m3}
Freund, Y. and Schapire, R. (1997).
\newblock A decision-theoretic generalization of on-line learning and an application to boosting.
\newblock {\em Journal of computer and system sciences}, 55:119--.

\bibitem[Galvao et~al., 2017]{b5}
Galvao, C.~B., Robles, L.~T., and Guerise, L.~C. (2017).
\newblock 20 years of port reform in brazil: Insights into the reform process.
\newblock {\em Research in transportation business and management}, 22:153--160.

\bibitem[Geurts et~al., 2006]{m5}
Geurts, P., Ernst, D., and Wehenkel, L. (2006).
\newblock Extremely randomized trees.
\newblock {\em Machine Learning}, 63:3--42.

\bibitem[Grinsztajn et~al., 2022]{Grinsztajn2022WhyDT}
Grinsztajn, L., Oyallon, E., and Varoquaux, G. (2022).
\newblock Why do tree-based models still outperform deep learning on typical tabular data?
\newblock In {\em Neural Information Processing Systems}.

\bibitem[Guo et~al., 2023]{lr1}
Guo, L., Zheng, J.-F., Liang, J., and Wang, S. (2023).
\newblock Column generation for the multi-port berth allocation problem with port cooperation stability.
\newblock {\em Transportation Research Part B: Methodological}, 171:3--28.

\bibitem[Hand and Yu, 2001]{m9}
Hand, D.~J. and Yu, K. (2001).
\newblock Idiot's bayes—not so stupid after all?
\newblock {\em International Statistical Review}, 69.

\bibitem[He et~al., 2015]{d2}
He, K., Zhang, X., Ren, S., and Sun, J. (2015).
\newblock Deep residual learning for image recognition.
\newblock {\em 2016 IEEE Conference on Computer Vision and Pattern Recognition (CVPR)}, pages 770--778.

\bibitem[Izaguirre et~al., 2020a]{p5}
Izaguirre, C., Losada, I.~J., Camus, P., Vigh, J.~L., and Stenek, V. (2020a).
\newblock Climate change risk to global port operations.
\newblock {\em Nature Climate Change}, 11:14--20.

\bibitem[Izaguirre et~al., 2020b]{Izaguirre2020ClimateCR}
Izaguirre, C., Losada, I.~J., Camus, P., Vigh, J.~L., and Stenek, V. (2020b).
\newblock Climate change risk to global port operations.
\newblock {\em Nature Climate Change}, 11:14--20.

\bibitem[James et~al., 2013]{m7}
James, G.~M., Witten, D.~M., Hastie, T.~J., and Tibshirani, R. (2013).
\newblock An introduction to statistical learning.
\newblock {\em Springer Texts in Statistics}.

\bibitem[Ke et~al., 2017]{m4}
Ke, G., Meng, Q., Finley, T., Wang, T., Chen, W., Ma, W., Ye, Q., and Liu, T.-Y. (2017).
\newblock Lightgbm: A highly efficient gradient boosting decision tree.
\newblock In Guyon, I., Luxburg, U.~V., Bengio, S., Wallach, H., Fergus, R., Vishwanathan, S., and Garnett, R., editors, {\em Advances in Neural Information Processing Systems}, volume~30. Curran Associates, Inc.

\bibitem[Kohavi, 1995]{Kohavi1995ASO}
Kohavi, R. (1995).
\newblock A study of cross-validation and bootstrap for accuracy estimation and model selection.
\newblock In {\em International Joint Conference on Artificial Intelligence}.

\bibitem[Kolley et~al., 2022a]{i2}
Kolley, L., R{\"u}ckert, N., Kastner, M., Jahn, C., and Fischer, K. (2022a).
\newblock Robust berth scheduling using machine learning for vessel arrival time prediction.
\newblock {\em Flexible Services and Manufacturing Journal}, 35:29--69.

\bibitem[Kolley et~al., 2022b]{h2}
Kolley, L., R{\"u}ckert, N., Kastner, M., Jahn, C., and Fischer, K. (2022b).
\newblock Robust berth scheduling using machine learning for vessel arrival time prediction.
\newblock {\em Flexible Services and Manufacturing Journal}, 35:29--69.

\bibitem[Lim et~al., 2019]{Lim2019PortSA}
Lim, S., Pettit, S., Abouarghoub, W., and Beresford, A. (2019).
\newblock Port sustainability and performance: A systematic literature review.
\newblock {\em Transportation Research Part D: Transport and Environment}.

\bibitem[L{\'o}pez-Berm{\'u}dez et~al., 2019]{p3}
L{\'o}pez-Berm{\'u}dez, B., Freire-Seoane, M.~J., and Gonz{\'a}lez-Laxe, F. (2019).
\newblock Efficiency and productivity of container terminals in brazilian ports (2008–2017).
\newblock {\em Utilities Policy}.

\bibitem[Louppe et~al., 2013]{m6}
Louppe, G., Wehenkel, L., Sutera, A., and Geurts, P. (2013).
\newblock Understanding variable importances in forests of randomized trees.
\newblock In {\em Neural Information Processing Systems}.

\bibitem[Lundberg and Lee, 2017]{shap}
Lundberg, S.~M. and Lee, S.-I. (2017).
\newblock A unified approach to interpreting model predictions.
\newblock In {\em Neural Information Processing Systems}.

\bibitem[Mekkaoui et~al., 2022]{i3}
Mekkaoui, S.~E., Benabbou, L., and Berrado, A. (2022).
\newblock Machine learning models for efficient port terminal operations: Case of vessels’ arrival times prediction.
\newblock {\em IFAC-PapersOnLine}.

\bibitem[Menze et~al., 2009]{fi}
Menze, B.~H., Kelm, B.~M., Masuch, R., Himmelreich, U., Bachert, P., Petrich, W., and Hamprecht, F.~A. (2009).
\newblock A comparison of random forest and its gini importance with standard chemometric methods for the feature selection and classification of spectral data.
\newblock {\em BMC Bioinformatics}, 10:213 -- 213.

\bibitem[Nguyen et~al., 2019]{p1}
Nguyen, T., Zhang, J., Zhou, L., and He, Y. (2019).
\newblock A data-driven optimization of large-scale dry port location using the hybrid approach of data mining and complex network theory.
\newblock {\em Transportation Research Part E: Logistics and Transportation Review}, 134.

\bibitem[Ogura et~al., 2021]{i1}
Ogura, T., Inoue, T., and Uchihira, N. (2021).
\newblock Prediction of arrival time of vessels considering future weather conditions.
\newblock {\em Applied Sciences}, 11:4410.

\bibitem[Reggiannini et~al., 2019]{Reggiannini2019RemoteSF}
Reggiannini, M., Righi, M., Tampucci, M., Duca, A.~L., Bacciu, C., Bedini, L., D'Errico, A., Paola, C.~D., Marchetti, A., Martinelli, M., Mercurio, C., Salerno, E., and Zizi, B. (2019).
\newblock Remote sensing for maritime prompt monitoring.
\newblock {\em Journal of Marine Science and Engineering}.

\bibitem[Rodrigues and Agra, 2022]{Rodrigues2022BerthAA}
Rodrigues, F. and Agra, A. (2022).
\newblock Berth allocation and quay crane assignment/scheduling problem under uncertainty: A survey.
\newblock {\em Eur. J. Oper. Res.}, 303:501--524.

\bibitem[Schmidhuber, 2014]{d3}
Schmidhuber, J. (2014).
\newblock Deep learning in neural networks: An overview.
\newblock {\em Neural networks : the official journal of the International Neural Network Society}, 61:85--117.

\bibitem[van Boetzelaer et~al., 2014]{Boetzelaer2013MastersTM}
van Boetzelaer, F.~B., van~den Boom, T., and Negenborn, R.~R. (2014).
\newblock Model predictive scheduling for container terminals.
\newblock {\em IFAC Proceedings Volumes}, 47:5091--5096.

\bibitem[Wang et~al., 2021]{Wang2021AnAO}
Wang, H., Liu, Z., Wang, X., Graham, T.~L., and Wang, J. (2021).
\newblock An analysis of factors affecting the severity of marine accidents.
\newblock {\em Reliab. Eng. Syst. Saf.}, 210:107513.

\bibitem[Weerasinghe et~al., 2023]{Weerasinghe2023OptimizingCT}
Weerasinghe, B.~A., Perera, H.~N., and Bai, X. (2023).
\newblock Optimizing container terminal operations: a systematic review of operations research applications.
\newblock {\em Maritime Economics \& Logistics}.

\bibitem[Zou and Hastie, 2005]{m11}
Zou, H. and Hastie, T.~J. (2005).
\newblock Addendum: Regularization and variable selection via the elastic net.
\newblock {\em Journal of the Royal Statistical Society: Series B (Statistical Methodology)}, 67.

\bibitem[Çağatay Iris and Lam, 2019]{p4}
Çağatay Iris and Lam, J. S.~L. (2019).
\newblock A review of energy efficiency in ports: Operational strategies, technologies and energy management systems.
\newblock {\em Renewable and Sustainable Energy Reviews}, 112:170--182.

\bibitem[Štepec et~al., 2020]{h3}
Štepec, D., Martinčič, T., Klein, F., Vladušič, D., and Costa, J.~P. (2020).
\newblock Machine learning based system for vessel turnaround time prediction.
\newblock In {\em 2020 21st IEEE International Conference on Mobile Data Management (MDM)}, pages 258--263.

\end{thebibliography}

\end{document}